\documentclass[numbers]{elsarticle}
\usepackage{lineno,hyperref}

\modulolinenumbers[5]
\usepackage{amsmath,amssymb,amsfonts}
\usepackage{algorithmic}
\usepackage{graphicx}
\usepackage{textcomp}
\usepackage{multirow}
\usepackage{xcolor}
\usepackage{natbib}

\definecolor{mycolor1}{rgb}{0.7,0.3,0.3}

\definecolor{mycolor2}{rgb}{0.0,0.0,1}

\newcommand{\Real}{\mathbb{R}}
\newcommand{\Natural}{\mathbb{N}}
\newcommand{\Volume}{\mathcal{V}}
\newcommand{\step}{\mathrm{Step}}

\newcommand{\bfx}{\boldsymbol{x}}
\newcommand{\bfu}{\boldsymbol{u}}
\newcommand{\bfq}{\boldsymbol{q}}

\newcommand{\bfz}{\boldsymbol{z}}
\newcommand{\bfw}{\boldsymbol{w}}
\newcommand{\bfs}{\boldsymbol{s}}
\newcommand{\bfy}{\boldsymbol{y}}

\newcommand{\Cov}{\mathrm{Cov}}

\newtheorem{thm}{Theorem}
\newtheorem{rem}{Remark}
\newtheorem{cor}{Corollary}
\newtheorem{lem}{Lemma}

\newtheorem{assume}{Assumption}
\newtheorem{alg}{Algorithm}

\begin{document}

\begin{frontmatter}

\title{Fast Construction of Correcting Ensembles for Legacy Artificial Intelligence Systems: Algorithms and a Case Study}

\author[Leic,Leti,UNN]{Ivan Yu. Tyukin\corref{cor1}\fnref{f1}}\ead{I.Tyukin@le.ac.uk}
\author[Leic,UNN]{Alexander N. Gorban}\ead{a.n.gorban@le.ac.uk}
\author[Leic]{Stephen Green}\ead{slg46@le.ac.uk}
\author[Toyota]{Danil Prokhorov}\ead{dvprokhorov@gmail.com}

\address[Leic]{University of Leicester, Department of Mathematics, University Road, Leicester, LE1 7RH, UK}
\address[Leti]{Department of Automation and
        Control Processes, St. Petersburg State University of
        Electrical Engineering, Prof. Popova str. 5, Saint-Petersburg, 197376, Russian Federation}
\address[UNN]{Lobachevsky State University of Nizhnij Novgorod, Prospekt Gagarina 23, Nizhnij Novgorod,603022, Russian Federation}
\address[Toyota]{Toyota R $\&$ D,  Ann-Arbor, MI, USA}
\fntext[f1]{The work was supported by the {Ministry of Education and Science} of Russia (Project No. 14.Y26.31.0022) and  Innovate UK  Knowledge Transfer Partnership grants KTP009890 and KTP010522.}

\begin{abstract}
This paper presents a {new approach} for {constructing} simple and computationally efficient improvements of generic Artificial Intelligence (AI) systems, including Multilayer and Deep Learning neural networks. The improvements are small network ensembles {added to} the existing AI architectures. Theoretical foundations of the {approach} are based on  stochastic separation theorems and the ideas of the concentration of measure. We show that, subject to mild technical assumptions on statistical properties of internal signals in the original AI system,  {the approach enables fast removal of the AI's errors} with probability close to one on the datasets which {may be} exponentially large in dimension. The {approach} is illustrated with numerical examples and a case study {of digits recognition in American Sign Language}.
\end{abstract}


\end{frontmatter}

\section{Introduction}

{Artificial Intelligence (AI) systems have risen dramatically from being the subject of niche academic and practical interests to a
commonly accepted and widely-spread facet of modern life.} Industrial giants such as Google, Amazon, IBM, and Microsoft offer a broad range of AI-based services, including intelligent image and sound processing and recognition.

{Deep Learning and related computational technologies \cite{Brahma2016}, \cite{ResNet} are currently perceived as state-of-the art tools for producing various AI systems. In visual recognition tasks, these systems deliver unprecedented accuracy \cite{Russ:2015} at  reasonable computational costs \cite{SqueezeNet:2016}.} Despite these advances, several fundamental challenges hinder further progress of these technologies.

{All data-driven AI systems make mistakes, regardless of how well an AI system is trained. Some examples} of these have already received global public attention \cite{Uber}, \cite{BBC_face_recognition_fails}. Mistakes may arise due to {uncertainty in empirical data,} data misrepresentation, and imprecise or inaccurate training. Conventional approaches {to reducing spurious}  errors include altering training data and improving design procedures   \cite{Kuznetsova:2015}, \cite{Misra:2015}, \cite{Prest:2012}, \cite{Zheng:2016}, {transfer learning \cite{chen2015net2net}, \cite{Pratt:ANIP:1992}, \cite{yosinski2014transferable}  and privileged learning \cite{vapnik2017knowledge}.}  These approaches invoke extensive training procedures. {The training itself, whilst} eradicating some errors, may introduce new errors by the very nature of the steps involved (e.g. {randomized sampling of} mini-batches, randomized training sets etc). 

{In this work, we propose a training approach and a technology whereby errors of the original legacy AI are removed, with high probability,} via incorporation of small {\it neuronal ensembles} and their cascades into the original AI's decision-making. {Ensembles methods and classifiers' cascades (see e.g.  \cite{freund1997decision}, \cite{gama2000cascade}, \cite{rokach2010ensemble}, \cite{wangcui2017}  and references therein) constitute a well-known  framework for improving performance of classifiers. Here we demonstrate that geometry} of {high-dimensional spaces, }  {in agreement with   ``blessing of dimensionality''   \cite{donoho2000high}, \cite{gorban2018blessing}, \cite{kainen1997utilizing}} {enables efficient} {construction of AI error correctors and their ensembles with guaranteed performance bounds.}

{The technology bears similarity with neurogenesis deep learning \cite{draelos2016neurogenesis}, classical cascade correlation \cite{fahlman1990cascade}, greedy approximation \cite{Barron}, and randomized methods for training neural networks \cite{scardapane2017randomness}, including deep stochastic configuration networks \cite{wang2017deep}, \cite{wang2017stochastic}.} The proposed technology, however, does not require computationally expensive training or pre-conditioning, and in some instances can be set up as a non-iterative procedure. In the latter case the {technology's} computational complexity scales at most linearly with the size of the training set.

 At the core of the technology is the concentration of measure phenomenon \cite{Gibbs1902}, \cite{Gorban:2007}, \cite{Gromov:1999}, \cite{GAFA:Gromov:2003}, \cite{Levi1951}, a view on the problem of learning in AI systems that is stemming from conventional probabilistic settings \cite{cucker2002mathematical}, \cite{Vapnik:2000}, and stochastic separation theorems \cite{GorbanTyukin:NN:2017}, \cite{gorban2018blessing}, \cite{GGGMT2018prob}. Main building blocks of the technology are simple threshold, perceptron-type \cite{Rosenblatt1962}  classifiers. The original AI system, however, {needs} not be a classifier itself. We show that, subject to mild assumptions on statistical properties {of signals} in the original AI system, small neuronal ensembles {consisting of} cascades of simple linear classifiers are an efficient tool for learning away spurious and systematic errors. These cascades can be used for learning new {tasks} too.

The paper is organized as follows: Section \ref{sec:statement} contains a formal statement of the problem, Section \ref{sec:result} presents main results: namely, the algorithms for constructing correcting ensembles (Section \ref{sec:algorithm}) preceded by their  mathematical justification (Section \ref{sec:math_preliminaries}).  Section \ref{sec:examples} presents a case study on American Sign Language recognition illustrating the concept, and Section \ref{sec:conclusion} concludes the paper. Proofs of theorems and other statements as well as additional technical results are provided {in \ref{sec:appendix_proofs} and \ref{sec:planes_vs_ellipses}}.

\section*{Notation}

The following notational agreements are used throughout the paper:
\begin{itemize}
\item $\Real$  denotes the field of real numbers;
\item $\Natural$  is the set of natural numbers;
\item $\Real^n$ stands for the $n$-dimensional real space; unless stated otherwise symbol $n$ is reserved to denote dimension of the underlying linear space;
\item let $\bfx,\bfy$ be elements of $\Real^n$, then $\langle \bfx,\bfy\rangle = \sum_{i=1}^n x_i y_i$ is their inner product;
\item let $\bfx\in\Real^n$, then $\|\bfx\|$ is the Euclidean norm of $\bfx$: $\|\bfx\|=\sqrt{\langle \bfx,\bfx\rangle}$;
\item $B_n(R)$ denotes a $n$-ball of radius $R$ centered at $0$: $B_n(R)=\{\bfx\in\Real^n | \ \|\bfx\|\leq R\}$;
\item if $\mathcal{S}$ is a finite set then $|\mathcal{S}|$ stands for its cardinality;
\item $\Volume(\Xi)$ is the Lebesgue volume of  $\Xi \subset \Real^n$;
\item if $X$ is a random variable then $E[X]$ is its expected value.
\end{itemize}

\section{Problem Statement}\label{sec:statement}

\begin{figure}
\centering
\includegraphics[width=\textwidth]{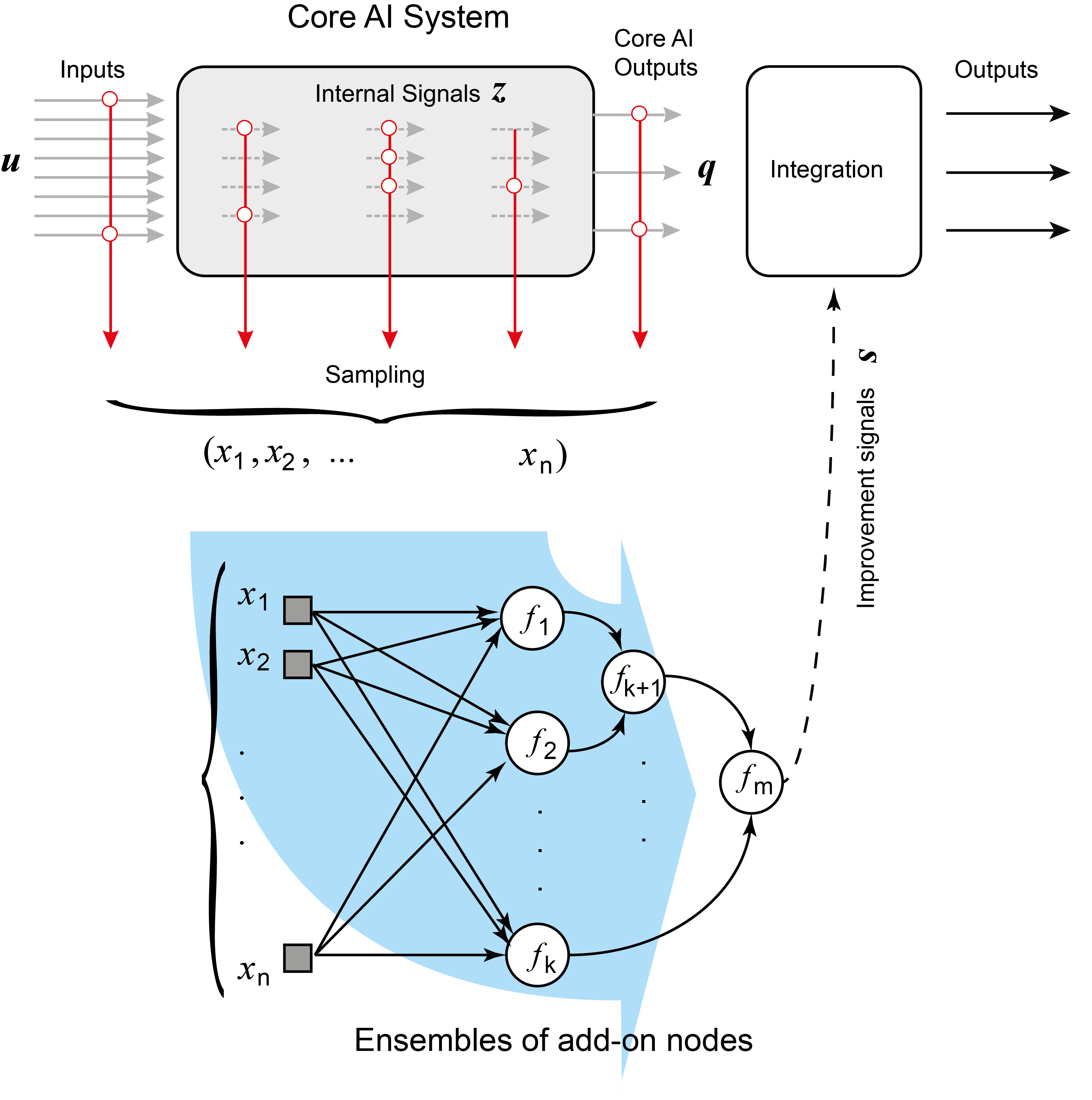}
\caption{Functional diagram of correcting neuronal ensembles for a generic Artificial Intelligence system.}\label{fig:diagram}
\end{figure}

 In what follows we suppose that an AI system is an operator mapping elements of its input set, $\mathcal{U}$, to the set of outputs, $\mathcal{Q}$. Examples of inputs $\bfu\in\mathcal{U}$ are images, temporal or spatiotemporal signals, and the outputs $\bfq\in\mathcal{Q}$ correspond to  labels, classes, or some quantitative characteristics of the inputs. Inputs $\bfu$, outputs $\bfq$, and internal variables $\bfz\in\mathcal{Z}$ of the system represent the system's state. The state itself may not be available for observation but some of its variables or relations may be accessed. In other words, we assume that there is a process which assigns an element of $\bfx\in\Real^n$ to the triple $(\bfu,\bfz,\bfq)$. A diagram illustrating the setup for a generic AI system is shown in Fig. \ref{fig:diagram}.

In addition to the core AI system we consider an add-on ensemble mapping samples $\bfx$ into auxiliary signals $\boldsymbol{s}\in\mathcal{S}$ (shown as {\it improvement signals} in Fig. \ref{fig:diagram}) which are then used to improve performance of the original  core AI. An example of such an improvement signal could be an indication that the current state of the core AI system represented by $\bfx$ corresponds to an erroneous decision of the core AI. At the integration stage, this information will alter the overall response.

With regards to the operations $f_i$ performed at the individual nodes (circles in the diagram on Fig. \ref{fig:diagram}), we will focus on the following relevant class
\begin{equation}\label{eq:nodes_class}
f_i (\bfx)  = f\left(\langle\bfw_i,\bfx\rangle - c_i\right),
\end{equation}
where the function $f:\Real\rightarrow\Real$ is piece-wise continuous, $\bfw_i\in\Real^n$, and $c_i\in\Real$. This class of functions is broad enough to enable the ensemble to function as a universal approximation device (e.g. by choosing the function $f$ in the class satisfying conditions discussed in \cite{Barron})
as well as to fit into a wide range of common AI architectures including decision trees, multilayer neural networks and perceptions, and deep learning convolutional neuronal networks.
Additional insight into ``isolation'' properties of elements (\ref{eq:nodes_class}) in comparison with more natural choices such as e.g. ellipsoids or balls (cf. \cite{Anderson2014}) is provided in \ref{sec:planes_vs_ellipses}.

Over a relevant period of time, the AI system generates a finite but large set of measurements $\bfx_i$. This set is assessed by an external supervisor and is partitioned into the union of the sets $\mathcal{M}$ and $\mathcal{Y}$:
\[
\mathcal{M}=\{\bfx_1,\dots,\bfx_{M}\},\ \mathcal{Y}=\{\bfx_{M+1},\dots,\bfx_{M+k}\}.
\]
The set $\mathcal{M}$ may contain measurements corresponding to expected operation of the AI, whereas elements from $\mathcal{Y}$ constitute core AI's performance singularities. These singularities may be both desired (related e.g. to ``important'' inputs $\bfu$) and undesired (related e.g. to errors of the AI). The function of the ensemble is to respond to these singularities selectively by producing improvement signals $\bfs$ in response to elements from the set $\mathcal{Y}$.

{A straightforward approach to derive such ensembles is to use standard optimization facilities such as e.g. error back-propagation and its various modifications, or use} other iterative procedures similar to greedy approximation \cite{Barron}, stochastic configuration networks \cite{wang2017stochastic} (cf. \cite{draelos2016neurogenesis}). This approach, however, requires computational resources and time that may not necessarily be available. Even in the simplest case of a single-element add-on system that is to separate the entire set  $\mathcal{X}$ from $\mathcal{Y}$, determining the best possible solution, in the form of e.g. support vector machines \cite{Vapnik:2000}, is non-trivial computationally. Indeed, theoretical worst-case estimates of computational complexity for determining parameters of the single-element system are of the order $O((M+k)^3)$ \cite{chapelle2007training}.

The question, therefore is: if there exists a computationally efficient procedure for determining ensembles or add-on networks such that
\begin{itemize}
\item[1)] they are able to ``improve'' performance of a generic AI system  with guaranteed probability, and
\item[2)] they consist of  elements (\ref{eq:nodes_class})?
\end{itemize}
Preferably, computational complexity of the procedure is to be linear or even sub-linear in $M+k$.

In the next sections we show that, subject to some mild technical assumptions on the way the sets $\mathcal{M}$ and $\mathcal{Y}$ are generated, a family of simple algorithms with the required characteristics can indeed be derived. These algorithms are motivated by Stochastic Separation Theorems \cite{GorbanTyukin:NN:2017}, \cite{gorban2018blessing}, \cite{GGGMT2018prob}. For consistency, adapted version of these theorems are presented and discussed in Section \ref{sec:math_preliminaries}. The algorithms themselves are presented and discussed in Section \ref{sec:algorithm}.

\section{Main Results}\label{sec:result}

\subsection{Mathematical Preliminaries}\label{sec:math_preliminaries}

Following standard assumptions (see e.g. \cite{cucker2002mathematical}, \cite{Vapnik:2000}), we suppose that all $\bfx$ are generated in accordance with some distribution, and the actual measurements $\bfx_i$ are samples from this distribution. For simplicity, we adopt a traditional setting in which all such samples be identically and independently distributed (i.i.d.) \cite{cucker2002mathematical}. With regards to the elements of $\bfx_i$, the following technical condition is assumed:
\begin{assume}\label{assume:product_measure} Elements $\bfx_i$ are random i.i.d. vectors drawn from a product measure distribution:
\begin{itemize}
\item[A1) ] their $x_{ij}$-th components are independent and bounded random variables $X_j$: $-1 \leq X_j\leq 1$,  $j=1,\dots,n$,
\item[A2) ] $E[X_j]=0$, and $E[X_j^2]=\sigma_j^2$.
\end{itemize}
\end{assume}
The distribution itself, however, is supposed to be unknown. Let
\[
R_0^2=\sum_{i=1}^n \sigma_i^2>0.
\]
Then the following result holds (cf. \cite{GorbanTyukin:NN:2017}).
\begin{thm}\label{thm:cube:1} Let $\bfx_i\in\mathcal{M}\cup\mathcal{Y}$ be i.i.d. random points  from the product distribution satisfying Assumption \ref{assume:product_measure}, $0< \delta <1$, $0<\varepsilon<1$ and $R_0>0$. Then
\begin{itemize}
\item[1) ] for any $i$,
\begin{equation}\label{eq:norms_cube}
\begin{split}
&P\left(1-\varepsilon  \leq \frac{\|\boldsymbol{x}_i\|^2}{R^2_0}\leq 1+\varepsilon \right)\geq 1- 2\exp \left(-\frac{2R_0^4 \varepsilon^2}{n} \right);\\
&P\left(\frac{\|\boldsymbol{x}_i\|^2}{R^2_0}\geq 1-\varepsilon\right)\geq 1- \exp \left(-\frac{2R_0^4 \varepsilon^2}{n} \right);
\end{split}
\end{equation}
\item[2) ] for any $i,j, \ i\neq j$,
\begin{equation}\label{eq:dot_product_cube}
\begin{split}
&P\left( \left\langle\frac{\bfx_i}{R_0}, \frac{\bfx_j}{R_0}\right\rangle < \delta \right)  \geq 1 - \exp \left(- \frac{R_0^4\delta^2}{2n} \right);\\
&P\left( \left\langle\frac{\bfx_i}{R_0}, \frac{\bfx_j}{R_0}\right\rangle >-\delta \right)  \geq 1 - \exp \left(- \frac{R_0^4\delta^2}{2n} \right);
\end{split}
\end{equation}
\item[3) ] for any given $\boldsymbol{y}\in[-1,1]^n$ and any $i$
\begin{equation}\label{eq:dot_product_cube_fixed}
\begin{split}
&P\left( \left\langle\frac{\bfx_i}{R_0}, \frac{\boldsymbol{y}}{R_0}\right\rangle < \delta \right)  \geq 1 - \exp \left(- \frac{R_0^4\delta^2}{2n} \right).
\end{split}
\end{equation}
\end{itemize}
\end{thm}
Proof of Theorem \ref{thm:cube:1} as well as those of other technical statements are provided in  \ref{sec:appendix_proofs}.

\begin{rem} \normalfont Notice that if $\sigma_i^2 > 0 $ then $R_0^2> n \min_i \{\sigma_i^2\}$. Hence the r.h.s. of (\ref{eq:norms_cube})--(\ref{eq:dot_product_cube_fixed}) become exponentially close to $1$ for $n$ large enough.
\end{rem}

The following Theorem is now immediate.
\begin{thm}[$1$-Element separation]\label{thm:cube:1-separation} Let elements of the set $\mathcal{M}\cup\mathcal{Y}$ be i.i.d. random points  from the product distribution satisfying Assumption \ref{assume:product_measure}, $0< \varepsilon <1$, and $R_0>0$. Consider $\bfx_{M+1}\in\mathcal{Y}$ and let
\begin{equation}\label{eq:1-functional}
\ell_1(\bfx)=\left\langle\frac{\bfx}{R_0},\frac{\bfx_{M+1}}{R_0}\right\rangle, \ h_1(\bfx)=\ell_1(\bfx) - 1 + \varepsilon.
\end{equation}
Then
\begin{equation}\label{eq:1-separation}
\begin{split}
&P\left( h_1(\bfx_{M+1})\geq 0  \ \mbox{and} \ h_1(\bfx_i)<0 \ \mbox{for all} \ \bfx_i\in\mathcal{M}\right)  \\
& \geq 1 - \exp \left(- \frac{2 R_0^4\varepsilon^2}{n} \right) - M \exp \left(- \frac{R_0^4(1-\varepsilon)^2}{2n} \right).
\end{split}
\end{equation}
\end{thm}
(The proof is provided in \ref{sec:appendix_proofs}.)

\begin{rem} \normalfont Theorem \ref{thm:cube:1-separation} not only establishes the fact that the set $\mathcal{M}$ can be separated away from $\mathcal{Y}$ by a linear functional with reasonably high probability. It also specifies the separating hyperplane, (\ref{eq:1-functional}), and provides an estimate from below of the probability of such an event,  (\ref{eq:1-separation}). The estimate, as a function of $n$, approaches $1$ exponentially fast. Note that the result is intrinsically related to the work \cite{Kurkova} on quasiorthogonal dimension of Euclidian spaces.
\end{rem}

Let us now move to the case when the set $\mathcal{Y}$ contains more than one element. Theorem \ref{thm:cube:k-separation} below summarizes the result.

\begin{thm}[$k$-Element separation. Case 1]\label{thm:cube:k-separation} Let elements of the set $\mathcal{M}\cup\mathcal{Y}$ be i.i.d. random points  from the product distribution satisfying Assumption \ref{assume:product_measure}, and $R_0>0$. Let, additionally,
\begin{equation}\label{eq:correlation_bound}
\left\langle \frac{\bfx_{i}}{R_0}, \frac{\bfx_{j}}{R_0}  \right\rangle \geq \beta
\end{equation}
for all $\bfx_{i},\bfx_{j}\in\mathcal{Y}$, $\bfx_{i}\neq \bfx_{j}$.  Pick
\[
0< \varepsilon <1, \ 1-\varepsilon+\beta(k-1)>0,
\]
and consider
\begin{equation}\label{eq:k-functional}
\begin{split}
\ell_k(\bfx)=&\left\langle\frac{\bfx}{R_0},\frac{\overline{\bfx}}{R_0}\right\rangle, \ \overline{\bfx}=\frac{1}{k}\sum_{i=1}^{k} \bfx_{M+i},\\
h_k(\bfx)=&\ell_k(\bfx) -\frac{1-\varepsilon + \beta (k-1)}{k}.
\end{split}
\end{equation}
Then
\begin{equation}\label{eq:k-separation}
\begin{split}
&P\left( h_k(\bfx_{j})\geq 0  \ \& \ h_k(\bfx_i)<0 \ \mbox{for all} \ \bfx_i\in\mathcal{M}, \ \bfx_j\in\mathcal{Y} \right)  \\
& \geq 1 - k\exp \left(-\frac{2R_0^4 \varepsilon^2}{n} \right) -  M \exp \left(-\frac{R_0^4 (1-\varepsilon+\beta (k-1))^2}{2 k^2 n} \right).
\end{split}
\end{equation}
\end{thm}
(The proof is provided in \ref{sec:appendix_proofs}.)

The value of $\beta$ in estimate (\ref{eq:correlation_bound}) may not necessarily be available a-priori. If this is the case then the following corollaries from Theorems  \ref{thm:cube:1-separation}  and \ref{thm:cube:k-separation} may be invoked.

\begin{cor}[$k$-Element separation. Case 1]\label{cor:k-separation} Let elements of the set $\mathcal{M}\cup\mathcal{Y}$ be i.i.d. random points  from the product distribution satisfying Assumption \ref{assume:product_measure}, and $R_0>0$. Pick
\[
0<\delta, \  0< \varepsilon <1, \ 1-\varepsilon-\delta(k-1)>0,
\]
and let
\begin{equation}\label{eq:k-functional:uncorrelated}
\begin{split}
\ell_k(\bfx)=&\left\langle\frac{\bfx}{R_0},\frac{\overline{\bfx}}{R_0}\right\rangle, \ \overline{\bfx}=\frac{1}{k}\sum_{i=1}^{k} \bfx_{M+i},\\
h_k(\bfx)=&\ell_k(\bfx) -\frac{1-\varepsilon - \delta (k-1)}{k}.
\end{split}
\end{equation}
Then
\begin{equation}\label{eq:k-separation:uncorrelated}
\begin{split}
&P\left( h_k(\bfx_{j})\geq 0  \ \& \ h_k(\bfx_i)<0 \ \mbox{for all} \ \bfx_i\in\mathcal{M}, \ \bfx_j\in\mathcal{Y} \right)  \\
& \geq 1 - k\exp \left(-\frac{2R_0^4 \varepsilon^2}{n} \right) - \frac{k(k-1)}{2} \exp \left(-\frac{R_0^4 \delta^2}{2n} \right) \\
& -  M \exp \left(-\frac{R_0^4 (1-\varepsilon-\delta (k-1))^2}{2 k^2 n} \right).
\end{split}
\end{equation}
\end{cor}

\begin{cor}[$k$-Element separation. Case 2]\label{cor:cube:k-separation:2} Let elements of the set $\mathcal{M}\cup\mathcal{Y}$ be i.i.d. random points  from the product distribution satisfying Assumption \ref{assume:product_measure}, $0< \varepsilon <1$, $0<\mu<1-\varepsilon$ and $R_0>0$. Pick $\bfx_{j}\in\mathcal{Y}$ and consider
\begin{equation}\label{eq:k-functional:2}
\begin{split}
\ell_k(\bfx)=&\left\langle\frac{\bfx}{R_0},\frac{\bfx_{j}}{R_0}\right\rangle, \ h_k(\bfx)=\ell_k(\bfx) - 1 + \varepsilon+\mu, \\
\Omega=&\left\{\bfx\in\Real^n \ \left| \left\langle \frac{\bfx_j}{R_0}, \frac{\bfx_j-\bfx}{R_0}\right\rangle \leq \mu\right. \right\}
\end{split}
\end{equation}
Then
\begin{equation}\label{eq:k-separation:2}
\begin{split}
&P\left( h_k(\bfx)\geq 0  \ \mbox{and} \ h_k(\bfx_i)<0 \ \mbox{for all} \ \bfx_i\in\mathcal{M}, \bfx\in\Omega\right)  \\
& \geq 1 - k\exp \left(- \frac{2 R_0^4\varepsilon^2}{n} \right) - M \exp \left(- \frac{R_0^4(1-\varepsilon-\mu)^2}{2n} \right).
\end{split}
\end{equation}
\end{cor}
Proofs of the corollaries are provided in \ref{sec:appendix_proofs}.

\begin{figure}
\begin{center}
\includegraphics[width=0.65\columnwidth]{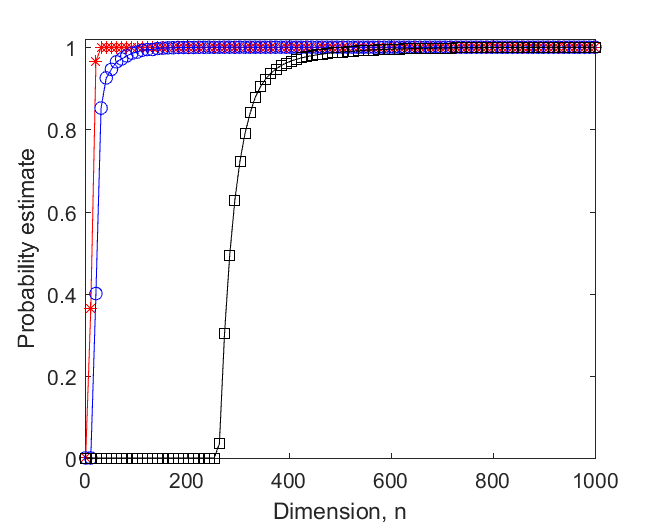}
\end{center}
\caption{The probability (frequency) that an element of the set $\mathcal{M}$ (formed by random i.i.d. vectors drawn from the n-cube $[-1,1]^n$) is  separable from the rest by hyperplanes  (\ref{eq:1-functional}) (and their variants) as a function of $n$;  $M=|\mathcal{M}|=10^4$. Red stars correspond to empirical frequencies of the events $A_i$: $\langle \bfx_i,\bfx_j\rangle < \|\bfx_i\|^2$ for all $\bfx_j\in\mathcal{M}$, $\bfx_i\neq \bfx_j$. Blue circles show frequencies of the events $B_i$: $\|\bfx_i\|^2/R_0^2 \geq 1-\varepsilon$ and $\langle \bfx_i/R_0,\bfx_j/R_0\rangle<1-\varepsilon$  for all $\bfx_j\in\mathcal{M}$, $\bfx_i\neq \bfx_j$, $\varepsilon=0.2$. Black squares show the right-hand side of (\ref{eq:cube_test}) for the same value of $\varepsilon=0.2$. }\label{fig:test_cube}
\end{figure}

Theorems \ref{thm:cube:1} -- \ref{thm:cube:k-separation} and Corollaries \ref{cor:k-separation}, \ref{cor:cube:k-separation:2} suggest that simple elements (\ref{eq:nodes_class}) with $f$ being mere threshold elements
\[
f(s)=\step(s)=\left\{\begin{array}{l} 1, \ s\geq 0 \\ 0, \ s<0 \end{array}\right.,
\]
posses remarkable selectivity. For example, according to Theorem \ref{thm:cube:1-separation}, the element
\[
f\left(\left\langle \bfx, \bfw\right\rangle - c\right), \ \bfw=\frac{\bfx_{M+1}}{R_0^2}, \ c=\frac{\|\bfx_{M+1}\|^2}{R_0^2}
\]
assigns ``$1$'' to $\bfx_{M+1}$ and ``$0$'' to all $\bfx_i\in\mathcal{M}$ with probability that is exponentially (in $n$) close to $1$. To illustrate this point, consider a simple test case with a set comprising of $10^4$ i.i.d. samples from the (uniform) product distribution in $[-1,1]^n$. For this distribution, $R_0=\sqrt{n/3}$, $|\mathcal{M}|=M=9999$, and estimate (\ref{eq:1-separation}) becomes:
\begin{equation}\label{eq:cube_test}
\begin{split}
&P\left( h_1(\bfx_{M+1})\geq 0  \ \mbox{and} \ h_1(\bfx_i)<0 \ \mbox{for all} \ \bfx_i\in\mathcal{M}\right)  \\
& \geq 1 - \exp \left(- \frac{2}{9} n\varepsilon^2\right) - M \exp \left(- \frac{1}{18}n(1-\varepsilon)^2 \right).
\end{split}
\end{equation}
The right-hand side rapidly approaches $1$ as $n$ grows. The estimate, however, could be rather conservative as is illustrated with Fig. \ref{fig:test_cube}.

A somewhat more relaxed approach could be to allow a small margin of error by determining a hyperplane separating the set $\mathcal{Y}$ from ``nearly all'' elements $\bfx\in\mathcal{M}$. An estimate of success for the latter case follows from Lemma \ref{lem:cascade} below

\begin{lem}\label{lem:cascade} Let elements of the set $\mathcal{M}$ be i.i.d. random points, and let
\[
\begin{split}
\ell(\bfx)=&\left\langle\bfx,\bfw\right\rangle, \ h(\bfx)=\ell(\bfx) - c
\end{split}
\]
and $0\leq p_{\ast}\leq 1$ be such that
\[
P\left( h(\bfx_i)\geq 0 \right)\leq p_{\ast}
\]
for an arbitrary element  $\bfx_i\in\mathcal{M}$. Then
\[
\begin{split}
& P\left(h(\bfx)\geq 0 \ \mbox{for at most} \ m \ \mbox{elements} \ \bfx\in\mathcal{M} \right)\geq \\
& \ \ \ \ (1-p_\ast)^{M} \exp\left(\frac{(M-m+1)p_{\ast}}{1-p_{\ast}}\right) \left(1-\frac{1}{m!}\left(\frac{(M-m+1)p_{\ast}}{(1-p_{\ast})}\right)^m \right).
\end{split}
\]
\end{lem}
(The proof is provided in \ref{sec:appendix_proofs}.)

\begin{figure}
\begin{center}
\includegraphics[width=0.65\columnwidth]{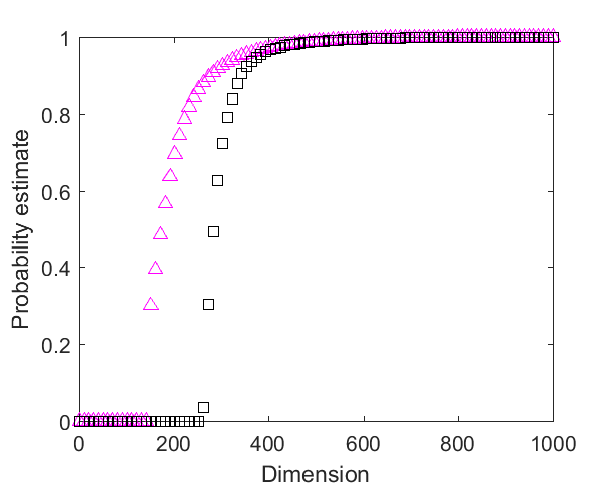}
\end{center}
\caption{Probability estimates (\ref{eq:cube_test}) (black squares) and (\ref{eq:1-separation_cascade}) (magenta triangles)  as functions of $n$ for the set $\mathcal{M}$ formed by random i.i.d. vectors drawn from the n-cube $[-1,1]^n$ and  $\varepsilon=0.2$.}\label{fig:test_cube:cascade}
\end{figure}

According to Lemma \ref{lem:cascade} and Theorem \ref{thm:cube:1},
\begin{equation}\label{eq:1-separation_cascade}
\begin{split}
&P\left( h_1(\bfx_{M+1})\geq 0  \ \mbox{and} \ h_1(\bfx_i)\geq 0 \ \mbox{for at most} \ n \  \bfx_i\in\mathcal{M}\right)  \\
& \geq  (1-p_\ast)^{M} \exp\left(\frac{(M-n+1)p_{\ast}}{1-p_{\ast}}\right) \left(1-\frac{1}{n!}\left(\frac{(M-n+1)p_{\ast}}{(1-p_{\ast})}\right)^n\right)\\
&  - \exp \left(- \frac{2 R_0^4\varepsilon^2}{n} \right), \  p_\ast=\exp \left(- \frac{R_0^4(1-\varepsilon)^2}{2n} \right).
\end{split}
\end{equation}
Observe, however, that $n+1$ random points (in general position) are linearly separable with probability $1$. Thus, with probability $1$, spurious $n$ points can  be separated away from $\bfx_{M+1}$ by a second hyperplane. Fig. \ref{fig:test_cube:cascade} compares separation probability estimate for such a pair with that of for $h_1$ (Theorem \ref{thm:cube:1-separation}, (\ref{eq:1-separation})).

%
%
%

\subsection{Fast AI up-training algorithms}\label{sec:algorithm}

Theorem \ref{thm:cube:k-separation} as well as Corollaries \ref{cor:k-separation}, \ref{cor:cube:k-separation:2} and Lemma \ref{lem:cascade} motivate a simple computational framework for construction of networks and cascades of elements (\ref{eq:nodes_class}). Following the setup presented in Section \ref{sec:statement}, recall that the data is sampled from the core AI system and is partitioned into the sets $\mathcal{M}$ and $\mathcal{Y}$. The latter set $\mathcal{Y}$ corresponds to singular events which are to be picked by the cascade. Their union, $\mathcal{S}=\mathcal{M}\cup\mathcal{Y}$, is the entire data that is available for the cascade construction.

We are now ready to proceed with the algorithms for fast construction of the correcting ensembles. The first algorithm is a recursion of which each iteration is a multi-step process. The algorithm is provided below.

\begin{alg}[Correcting Ensembles]\label{alg:algorithm_1}\normalfont
Initialization: set $\mathcal{M}^{1}$ $=$ $\mathcal{M}$, $\mathcal{Y}^{1}=\mathcal{Y}$, and $\mathcal{S}^{1}=\mathcal{S}$, define a list or a model of {\it correcting actions} -- formalized alternations of the core AI in response to an error/error type.
\begin{enumerate}

\item[] {\it Input to the $i$-th iteration:} Sets $\mathcal{M}^{i}$, $\mathcal{Y}^{i}$, and $\mathcal{S}^{i}=\mathcal{M}^{i}\cup\mathcal{Y}^{i}$; the number of clusters, $p$, and the value of filtering threshold, $\theta$.

\item {\it Centering}. All current data available is centered. The centered sets are denoted as $\mathcal{S}_c^{i}$ and $\mathcal{Y}_c^{i}$ and are formed by subtracting the mean $\overline{\bfx}(\mathcal{S}^i)$ from the elements of $\mathcal{S}^i$ and $\mathcal{Y}^i$, respectively:
$$\mathcal{S}_c^i = \{\bfx \in \mathbb{R}^n | \bfx = \xi - \overline{\bfx}(\mathcal{S}^i)\text{, }\xi \in \mathcal{S}^i \}$$
$$\mathcal{Y}_c^i = \{\bfx \in \mathbb{R}^n | \bfx = \xi - \overline{\bfx}(\mathcal{S}^i)\text{, }\xi \in \mathcal{Y}^i \}.$$
\item {\it Regularization}. The covariance matrix, $\Cov(\mathcal{S}^i)$, of $\mathcal{S}^i$ is calculated along with the corresponding eigenvalues and eigenvectors. New regularized sets $\mathcal{S}_r$, $\mathcal{Y}_r$ are produced as follows. All eigenvectors $h_1,h_2,\dots, h_m$ that correspond to the eigenvalues $\lambda_1,\lambda_2,\dots, \lambda_m$ which are above a given threshold are combined into a single matrix $H=(h_1 | h_2 | \cdots | h_m)$. The threshold could be chosen on the basis of the Kaiser-Guttman test \cite{Jackson:1993} or otherwise (e.g. by keeping the ratio of the maximal to the minimal in the set of retained eigenvalues within a given bound). The new sets are defined as:
$$\mathcal{S}_r^i = \{\bfx \in \Real^m | \bfx = H^T\xi\text{, }\xi \in \mathcal{S}_c^i\}$$
$$\mathcal{Y}_r^i = \{\bfx \in \Real^m | \bfx = H^T\xi\text{, }\xi \in \mathcal{Y}_c^i\}.$$
\item {\it Whitening}. The two sets then undergo a whitening coordinate transformation ensuring that the covariance matrix of the transformed data is the identity matrix:
$$\mathcal{S}_w^i = \{\bfx \in \Real^m | \bfx = W\xi\text{, }\xi \in \mathcal{S}_r^i, \ W=\Cov(\mathcal{S}_r^i)^{-\frac{1}{2}}\}$$
$$\mathcal{Y}_w^i = \{\bfx \in \Real^m | \bfx = W\xi\text{, }\xi \in \mathcal{Y}_r^i, \ W=\Cov(\mathcal{S}_r^i)^{-\frac{1}{2}}\}.$$
\item {\it Projection (optional)}. Project elements of $\mathcal{S}_w^i$, $\mathcal{Y}_w^i$ onto the unit sphere by scaling them to the unit length: $\bfx\mapsto \varphi(\bfx), \ \varphi(\bfx)=\bfx/\|\bfx\|$.

\item {\it Training: Clustering.} The set $\mathcal{Y}_w^i$  (the set of errors) is then partitioned into $p$ clusters $\mathcal{Y}_{w,1}^i,\dots,\mathcal{Y}_{w,p}^i$ that's elements are pairwise positively correlated.
\item {\it Training: Nodes creation and aggregation.} For each $\mathcal{Y}_{w,j}^i$, $j=1,\dots,p$ and its complement $\mathcal{S}_w^i \setminus \mathcal{Y}_{w\text{, }j}^i$ we construct the following separating hyperplanes:
\[
\begin{split}
h_j(\bfx)=& \ell_j(\bfx)-c_j, \\
\ell_j(\bfx) =& \left\langle\frac{\bfw_j}{\|\bfw_j\|}\text{, }\bfx\right\rangle, \ c_j= \min_{\xi \in \mathcal{Y}_{w\text{, }j}^i}\left\langle\frac{\bfw_j}{\|\bfw_j\|}\text{, }\xi\right\rangle\\
\bfw_j =& (\Cov(\mathcal{S}_w^{i} \setminus \mathcal{Y}_{w\text{,}j}^i) + \Cov(\mathcal{Y}_{w\text{,}j}^i))^{-1} (\overline{\bfx}(\mathcal{Y}_{w\text{,}j}^i) - \overline{\bfx}(\mathcal{S}_w^i \setminus \mathcal{Y}_{w\text{,}j}^i)).
\end{split}
\]
Retain only those hyperplanes for which $c_j>\theta$. For each retained hyperplane, create a corresponding element (\ref{eq:nodes_class})
\begin{equation}\label{eq:node:type_1}
f_j(\bfx)=f\left(\left\langle W H^{T} (\bfx - \overline{\bfx}(\mathcal{S}^i)),\frac{\bfw_j}{\|\bfw_j\|}\right\rangle-c_j\right),
\end{equation}
with $f$ being a function that satisfies: $f(s)> 0$ for all $s\geq 0$ and $f(s)\leq 0$ for all $s < 0$, and add it to the ensemble. If optional step 4 was used then the functional definition of $f_j$ becomes:
\begin{equation}\label{eq:node:type_2}
f_j(\bfx)=f\left(\left\langle \varphi\left(W H^{T} (\bfx - \overline{\bfx}(\mathcal{S}^i))\right),\frac{\bfw_j}{\|\bfw_j\|}\right\rangle-c_j\right),
\end{equation}
where $\varphi$ is the mapping implementing projection onto the unit sphere.

\item {\it Integration/Deployment.} Any $\bfx$ that is generated by the original core AI is  put through the ensemble of $f_j(\cdot)$. If for some $\bfx$ any of the values of $f_j(\bfx) > 0$ then  a {\it correcting action} is performed on the core AI. The action is dependent on the purpose of correction as well as on the problem at hand. It could include label swapping, signalling an alarm, ignoring/not reporting etc. The combined system becomes new core AI.

\item{\it Testing.} Assess performance of new core AI on a relevant data set. If needed, generate new sets $\mathcal{M}^{i+1}$, $\mathcal{Y}^{i+1}$,  $\mathcal{S}^{i+1}$ (with possibly different error types and definitions), and repeat the procedure.
\end{enumerate}
\end{alg}

\begin{figure}
\centering
\includegraphics[width=0.4\columnwidth]{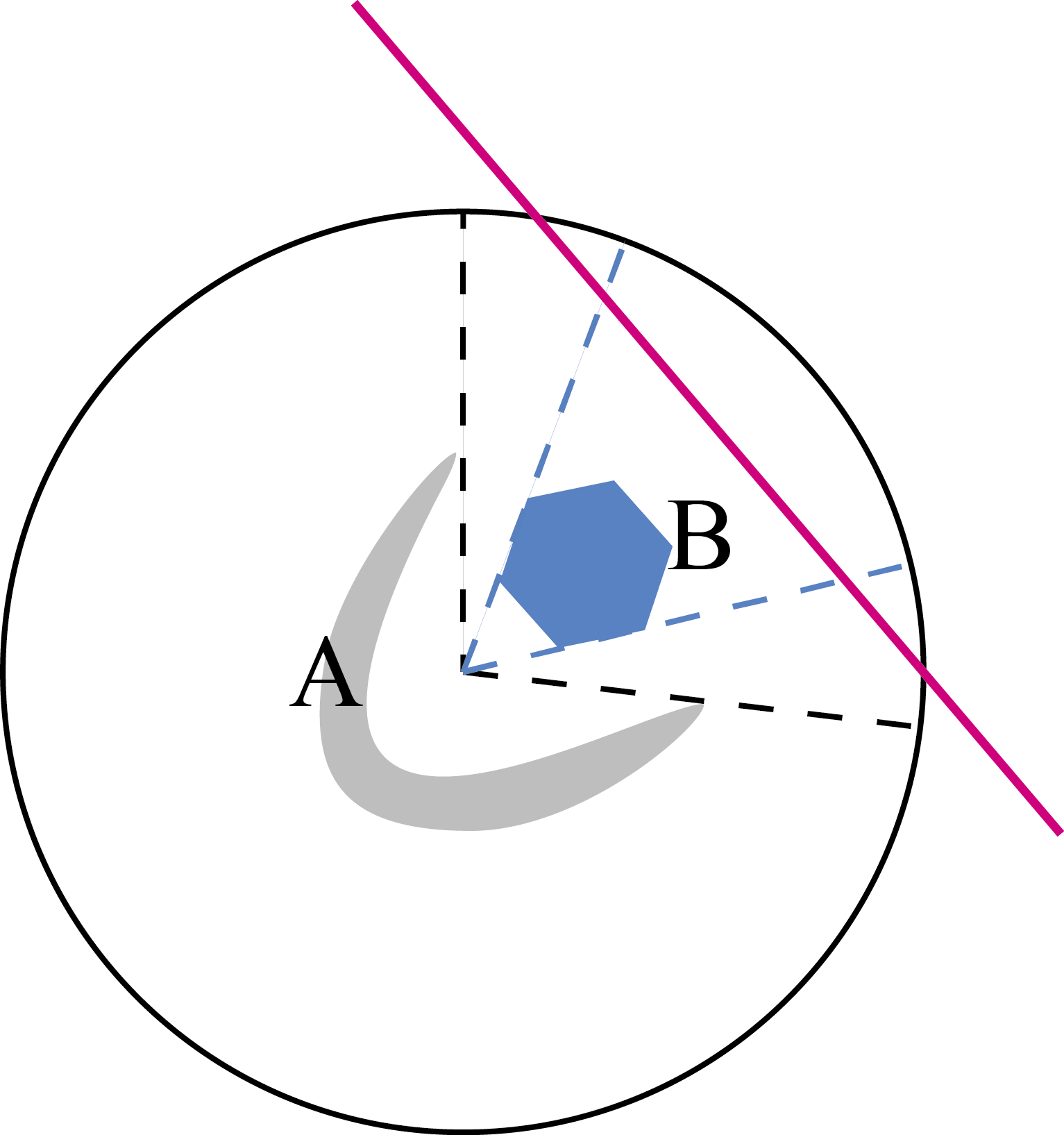}
\caption{Linear separability via projection onto a sphere. Sets $A$ (shown as a shaded gray domain) and $B$ (depicted as a blue filled hexagon) are not linearly separable in the original coordinates. Their projections onto the sphere, however, are separable (the separating hyperplane is shown in red).}\label{fig:spherical_proj}
\end{figure}

Steps 1--3 of the algorithm are standard pre-processing routines. In the context of Theorems \ref{thm:cube:1}--\ref{thm:cube:k-separation}, step 1 aims to ensure that the first part of A2) in Assumption  \ref{assume:product_measure} holds, and step 2, in addition to regularization, results in that all components of $\bfx_i$ are uncorrelated (a necessary condition for part  A1) of Assumption \ref{assume:product_measure} to hold). Whitening transformation, step 3, normalizes the data in the sense that $\sigma_i^2=1$ for all $i$ relevant.  Step 4 (optional) is introduced to account for data irregularities and clustering that may negatively affect data separability. An illustration of potential utility of this step is shown in Fig. \ref{fig:spherical_proj}.
Note that individual components of the data vectors may no longer satisfy the independence assumption. Nevertheless, if the data is reasonably distributed, different versions of separation theorems may apply to this case too \cite{gorban2018blessing}, \cite{GGGMT2018prob}. In Section \ref{sec:examples} we illustrate the effect of this step in an example application.

Step 5, clustering, is motivated by Theorem \ref{thm:cube:k-separation} and Corollaries {\ref{cor:k-separation}}, \ref{cor:cube:k-separation:2}  suggesting that the rate of success in isolating multiple points from the background set $\mathcal{M}$ increases when these multiple points are positively correlated or are spatially close to each other.

Step 6 is a version of (\ref{eq:k-functional}), albeit with a different normalization and a slight perturbation of weights. The choice of this particular normalization is motivated by the experiments shown in Fig. \ref{fig:test_cube}. As for the choice of weights, these hyperplanes are standard Fisher discriminants. Yet, they are not far from (\ref{eq:k-functional}). In view of the previous steps,  $\Cov(\mathcal{S}_w^{i} \setminus \mathcal{Y}_{w\text{,}j}^i) + \Cov(\mathcal{Y}_{w\text{,}j}^i)$ is diagonally dominated and close to the identity matrix. When $|\mathcal{S}_w^i|\gg |\mathcal{Y}_{w\text{,}j}^i|$, term $\overline{\bfx}(\mathcal{S}_w^i \setminus \mathcal{Y}_{w\text{,}j}^i)$ is nearly zero, and hence $\bfw_j$ is approximately at the centroid of the cluster. Filtering of the nodes is necessitated by the concentration of measure effects as exemplified by Theorem \ref{thm:cube:1}.

Functions $f$ in step 7 can be implemented using a range of ReLU, threshold, $\tanh(\cdot)$, sigmoidal functions etc.; these are available in the majority of standard core AI systems.

Computational complexity of each step in the recursion, except for step 5, is at most $M+k$. Complexity of the clustering step may generally be superpolynomial (worst case) as is e.g. the case for standard k-means clustering \cite{arthur2006slow}. If, however, sub-optimal solutions are accepted \cite{Hartigan:1979} then the complexity of this step scales linearly with $M+k$. Further speed-ups are possible through randomized procedures such as e.g. in k-means++ \cite{arthur2007k}.

Algorithm \ref{alg:algorithm_1} is based primarily on the intuition and rationale stemming from Theorem  \ref{thm:cube:1-separation}, \ref{thm:cube:k-separation} and their corollaries. It can be modified to take advantage of the possibility offered by Lemma \ref{lem:cascade} and the subsequent discussion. The modification is that  each node added in step 6 is assessed and ``corrected'' by an additional hyperplane if required. The modified algorithm is summarized as Algorithm \ref{alg:algorithm_2} below.

\begin{alg}[With cascaded pairs]\label{alg:algorithm_2}\normalfont
Initialization: set $\mathcal{M}^{1}=\mathcal{M}$, $\mathcal{Y}^{1}=\mathcal{Y}$, and $\mathcal{S}^{1}=\mathcal{S}$, define a list or a model of {\it correcting actions} -- formalized alternations of the core AI in response in response to an error/error type.
\begin{enumerate}

\item[] {\it Input to the $i$-th iteration:} Sets $\mathcal{M}^{i}$, $\mathcal{Y}^{i}$, and $\mathcal{S}^{i}=\mathcal{M}^{i}\cup\mathcal{Y}^{i}$; the number of clusters, $p$, and the value of filtering threshold, $\theta$.

\item[1-5.]  As in Algorithm \ref{alg:algorithm_1}.
\item[6.] {\it Training: Nodes creation and aggregation.} For each $\mathcal{Y}_{w,j}^i$, $j=1,\dots,p$ and its complement $\mathcal{S}_w^i \setminus \mathcal{Y}_{w\text{, }j}^i$  construct the following separating hyperplanes:
\[
\begin{split}
h_j(\bfx)=& \ell_j(\bfx)-c_j, \\
\ell_j(\bfx) =& \left\langle\frac{\bfw_j}{\|\bfw_j\|}\text{, }\bfx\right\rangle, \ c_j= \min_{\xi \in \mathcal{Y}_{w\text{, }j}^i}\left\langle\frac{\bfw_j}{\|\bfw_j\|}\text{, }\xi\right\rangle\\
\bfw_j =& (\Cov(\mathcal{S}_w^{i} \setminus \mathcal{Y}_{w\text{,}j}^i) + \Cov(\mathcal{Y}_{w\text{,}j}^i))^{-1} (\overline{\bfx}(\mathcal{Y}_{w\text{,}j}^i) - \overline{\bfx}(\mathcal{S}_w^i \setminus \mathcal{Y}_{w\text{,}j}^i)).
\end{split}
\]
Retain only those hyperplanes for which $c_j>\theta$. For each retained hyperplane, create a corresponding element $f_j(\cdot)$ in accordance to (\ref{eq:node:type_1}) or, if step 4 was used, then (\ref{eq:node:type_2}). For each retained hyperplane,
\begin{itemize}
\item determine the complementary set $\mathcal{C}_{j}$ comprised of elements $\bfx\in  \mathcal{S}_w^{i}\setminus \mathcal{Y}_w^{i}$ for which $h_j(\bfx)\geq 0$ (the set of points that are accidentally picked up by the $f_j$ ``by mistake'')
\item project elements of the set $\mathcal{C}_{j}\cup \mathcal{Y}_{w\text{,}j}^i$ orthogonally onto the hyperplane $h_j(\bfx)= \ell_j(\bfx)-c_j$ as
\[
\bfx \mapsto \left(I-\frac{1}{\|\bfw_j\|^2}\bfw_j \bfw_j^T\right)\bfx + c_j\frac{\bfw_j}{\|\bfw_j\|}=P(\bfw_j)\bfx + b(\bfw_j,c_j);
\]
\item determine a hyperplane
\[
h_{j,2}(\bfx)= \ell_{j,2}(\bfx)-c_{j,2}, \ \ell_{j,2}(\bfx)=\langle \bfw_{j,2}, \bfx\rangle;
\]
separating projections of $\mathcal{C}_{j}$ from that of $\mathcal{Y}_{w\text{,}j}^i$ so that $h_{j,2}(\bfx) < 0$ for all projections from $\mathcal{C}_{j}$ and $h_{j,2}(\bfx)\geq 0$ for all projections from $\mathcal{Y}_{w\text{,}j}^i$. If no such planes exist, use linear Fisher discriminant or any other relevant computational procedure
\item create a node
\[
f^{\perp}_j(\bfx)=f\left(\left\langle P (\bfw_j)WH^T(\bfx - \overline{\bfx}(\mathcal{S}^i)) + b(\bfw_j,c_j),\bfw_{j,2}\right\rangle-c_{j,2}\right)
\]
or, in case step 4 was used
\[
f^{\perp}_j(\bfx)=f\left(\left\langle P (\bfw_j)\varphi(WH^T(\bfx - \overline{\bfx}(\mathcal{S}^i))) + b(\bfw_j,c_j),\bfw_{j,2}\right\rangle-c_{j,2}\right);
\]
\item create the pair's response node, $f_j^{\mathrm{c}}$, so that a non-negative response is generated only when both nodes produce a non-negative response
\[
\begin{split}
f_j^{\mathrm{c}}(\bfx)=& f\big( \step(f_j(\bfx))+\step(f^{\perp}_j(\bfx)) - 2 \big);
\end{split}
\]
 \item add the combined  $f_j^{\mathrm{c}}(\cdot)$ to the ensemble.
\end{itemize}

\item[7. ] {\it Integration/Deployment.} Any $\bfx$ that is generated by the original core AI is  put through the ensemble of $f_j^{\mathrm{c}}(\cdot)$. If for some $\bfx$ any of the values of $f_j^{\mathrm{c}}(\bfx)> 0 $ then  a {\it correcting action} is performed on the core AI. The action is dependent on the purpose of correction as well as on the problem at hand. It could include label swapping, signalling an alarm, ignoring/not reporting etc. The combined system becomes new core AI.

\item[8. ] {\it Testing.} Assess performance of new core AI on a relevant data set. If needed, generate new sets $\mathcal{M}^{i+1}$, $\mathcal{Y}^{i+1}$,  $\mathcal{S}^{i+1}$ (with possibly different error types and definitions), and repeat the procedure.

\end{enumerate}
\end{alg}

In the next section we illustrate how the proposed algorithms work in a case study example involving a core AI in the form of a reasonably large convolutional network trained on a moderate-size dataset.

\section{Distinguishing the Ten Digits in American Sign Language: a Case Study}\label{sec:examples}

\subsection{Setup and Datasets}
\begin{figure}[h]
\centering
\includegraphics[width=0.9\textwidth]{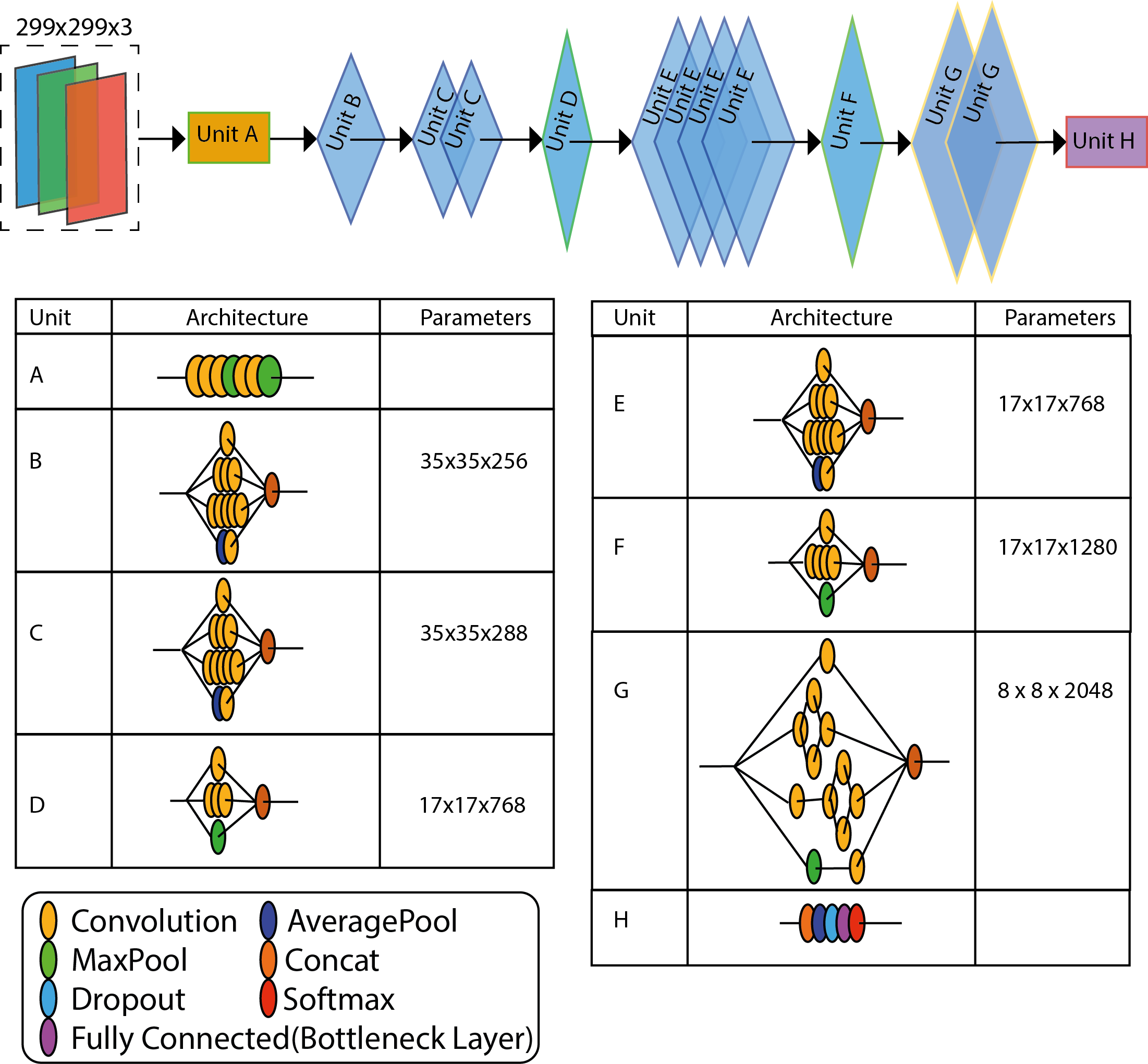}
\caption{Architecture of the adopted Inception deep neural network model.}\label{fig:Inception}
\end{figure}

\begin{figure}[h]
\centering
\includegraphics[scale=0.0167]{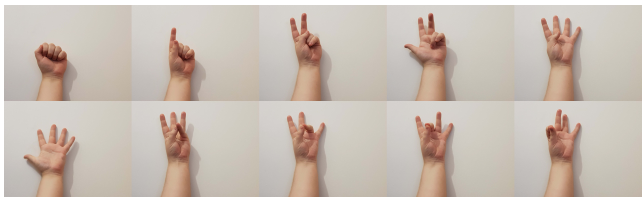}
\caption{Examples of the sort of images that appear in the current model's data set of the American Sign Language single hand positions for 0 (top left) to 9 (bottom right)}\label{fig:signs}
\end{figure}

 In this case study we investigated and tested the approach on a challenging problem of gestures recognition in the framework of distinguishing ten digits in American  Sign Language. To apply the approach a core AI had to be generated first. {As our core AI we picked a version of Inception deep neural network model \cite{szegedy2015going} whose architecture is shown in Fig. \ref{fig:Inception}.} {The model was trained}\footnote{https://www.tensorflow.org/tutorials/image$\text{\_}$retraining} on ten sets of images that correspond to the American Sign Language pictures for 0-9 (see Fig. \ref{fig:signs}).
{Each set contained 1000 unique images consisting of profile shots of the person’s hand, along with 3/4 profiles and shots from above and below.}

The states $\bfx_i$ are the vectors containing the values of pre-softmax layer bottlenecks of size $n$ for however many neurons are in the penultimate layer. {Schematically the network's layer whose outputs are $\bfx_i$ is shown in the Diagram in Fig. \ref{fig:Inception} as the fully connected (bottleneck layer) in Unit H.} Elements of this set that gave incorrect readings are noted and copied into the set $\mathcal{Y}$.

\subsection{Experiments and results} Once the network was trained, additional 10000 images of the same ratio were evaluated  using the trained system. For these experiments the classification decision rule was to return a gesture number that corresponds to the network output with the highest score (winner-takes-all) if the highest score exceeds a given threshold, $\gamma\in[0,1]$. Ties are broken arbitrarily at random. If the highest score is smaller than or equal to  the threshold then no responses are returned.

To assess performance of the resulting classifier, we used the following indicators as functions of threshold $\gamma$:
\begin{equation}\label{eq:performance_indicators}
\begin{split}
&\mbox{True Positive Rate }(\gamma)=\frac{TP (\gamma)}{FN(\gamma) + TP(\gamma)},\\
&\mbox{Missclassification Rate }(\gamma)= \frac{FP (\gamma)}{FP(0)},
\end{split}
\end{equation}
where $TP(\gamma)$ is the number of true positives, $FN(\gamma)$ is the number of false negatives, and $FP(\gamma)$ is the number of false positives. The definitions of these variables are provided in Table \ref{tab:errors_definition}.
\begin{table}
\small
\begin{center}
\caption{Error types in the system. Stars mark instances/events that are not accounted for.}\label{tab:errors_definition}
\begin{tabular}{|c|c|c|}
\hline
Presence of & System's & Error\\
a gesture & response & Type\\
\hline
\multirow{3}{*}{Yes} & \multicolumn{1}{c|}{ Correctly classified} & %
\multicolumn{1}{c|}{True Positive} \\
\cline{2-3}
& \multicolumn{1}{c|}{Incorrectly classified} & \multicolumn{1}{c|}{False Positive} \\
\cline{2-3}
& \multicolumn{1}{c|}{Not reported} & \multicolumn{1}{c|}{False Negative}\\
\hline
\multirow{2}{*}{No} & \multicolumn{1}{c|}{Reported} & %
\multicolumn{1}{c|}{False Positive*} \\
\cline{2-3}
& \multicolumn{1}{c|}{Not reported} & \multicolumn{1}{c|}{True Negative*} \\
\hline
\end{tabular}
\end{center}
\end{table}

In our experiments we did not add any negatives to the test set, as the original focus was to illustrate how the approach may cope with misclassification errors. This implies that the number of True Negatives is $0$ for any threshold $\gamma\in[0,1]$ and, consequently, the rate of false positives is always $1$. Therefore, instead of using traditional ROC curves showing the rate of true positives against the rate of false positives, we employed a different family of curves in which the rate of false positives is replaced with the rate of misclassification as defined by (\ref{eq:performance_indicators}).

\begin{figure}
\centering
\includegraphics[width=0.5\columnwidth]{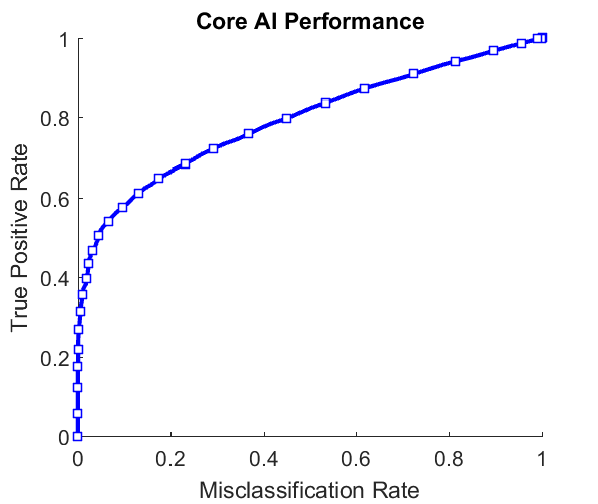}
\caption{Performance curve for the Core AI system (Inception) on the testing set of 10000 images. Threshold $\gamma$ was varying from $1$ (bottom left corner) to $0$ (top right corner)}\label{fig:ROC_core}
\end{figure}

A corresponding performance curve for this Core AI system  is shown in Fig. \ref{fig:ROC_core}.
At  $\gamma=0$ the result was an 82.4\% success rate for the adapted algorithm. At this end of the  curve, the observed performance was comparable/similar to that reported in e.g. \cite{bheda2017using} (see also references therein). The numbers of errors per each gesture in the trained system are shown in Table \ref{tab:3}.
\begin{table}
\begin{center}
\caption{Errors per gesture (misclassifications). Top row corresponds to the gesture number, the bottom row indicates the number of errors for each gesture.} \label{tab:3}
\small{
\begin{tabular}{|c|c|c|c|c|c|c|c|c|c|}
\hline
0 & 1 & 2 & 3 & 4 & 5 & 6 & 7 & 8 & 9 \\
\hline
10 & 52 & 235 & 62 & 410 & 80 & 269 & 327 & 207 & 108 \\
\hline
\end{tabular}}
\end{center}
\end{table}
The variance of errors is mostly consistent among the ten classes with very few errors for the ``0'' gesture, likely  due to its unique shape among the classes.

Once the errors were isolated, we used  Algorithms \ref{alg:algorithm_1} and \ref{alg:algorithm_2} to construct correcting ensembles improving the original core AI. To train the  ensembles, the testing data set of 10000 images that has been used to assess performance of Inception was split into two non-overlapping subsets. The first subset, comprised of 6592 records of data points corresponding to correct responses and 1408 records corresponding to errors, was used to train the correcting ensembles. This subset was the ensemble's training set, and it accounted for 80$\%$ of the data. The second subset, the ensemble's testing set, combining 1648 data points of correct responses and 352 elements labelled as errors was used to test the ensemble.

\begin{figure}
\begin{center}
\includegraphics[width=0.5\columnwidth]{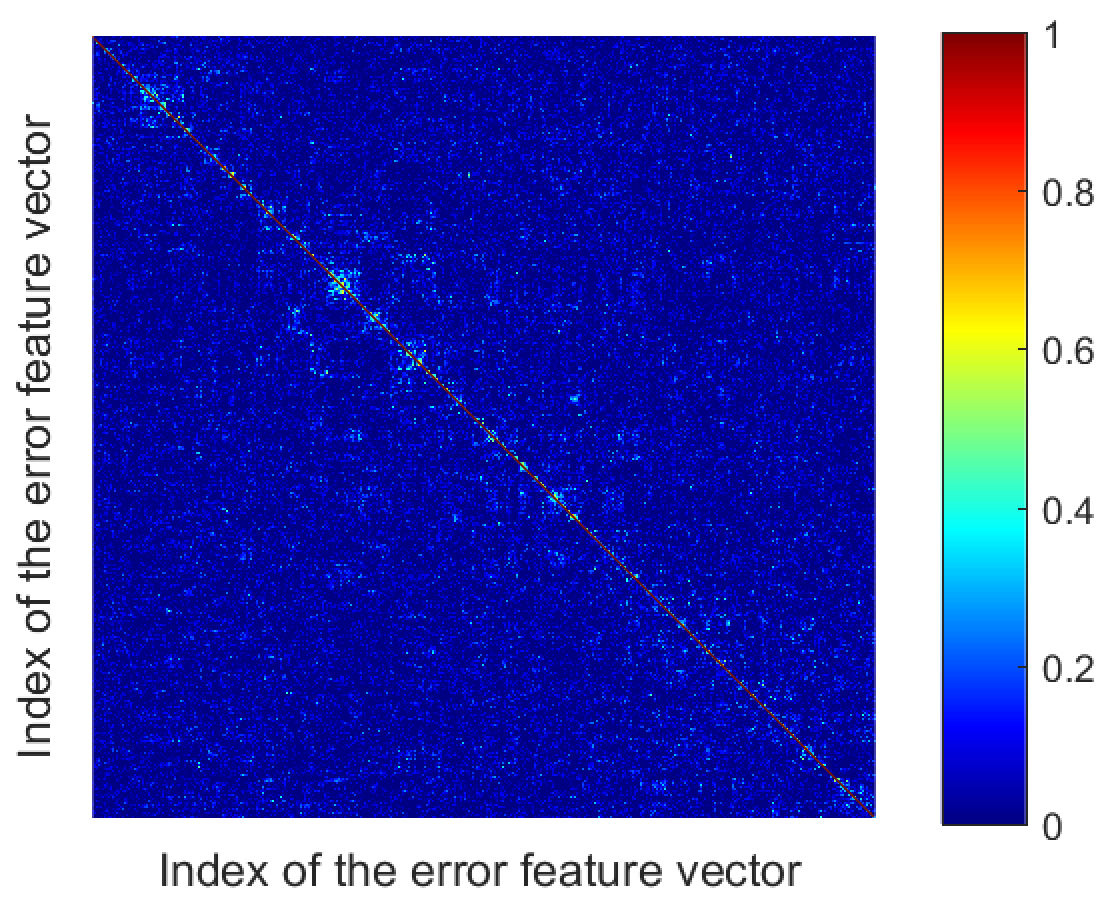}
\end{center}
\caption{Values of  $|\langle\bfx_i/\|\bfx_i\|,\bfx_j/\|\bfx_j\|\rangle|$ (color-coded) for the data points labelled as errors.}\label{fig:correlations_data}
\end{figure}

Both algorithms have been run on the first subset, the ensemble's training set. For simplicity of comparison and evaluation, we iterated the algorithms only once (i.e. did not build cascades of ensembles). In the regularization step, step 2, we used Kaiser-Guttman test. This returned $174$ principal components reducing the original dimensionality more than 10 times. After the whitening transformation, step 3, we assessed the values of $|\langle\bfx_i/\|\bfx_i\|,\bfx_j/\|\bfx_j\|\rangle|$ (shown in Fig. \ref{fig:correlations_data}).
According to Fig. \ref{fig:correlations_data}, data points labelled as errors are largely orthogonal to each other apart from few modestly-sized groups.

The number of clusters, parameter $p$ in step 5, was varied from 2 to 1408 in regular increments. As a clustering algorithm we used standard k-means routine (k-means ++) supplied with MATLAB 2016a. For each value of $p$, we run the k-means algorithm 10 times. For each clustering pass we constructed the corresponding nodes $f_j$ (or their pairs for Algorithm \ref{alg:algorithm_2}) as prescribed in step 6, and combined them into a single correcting ensemble in accordance with step 7.

\begin{figure}
\begin{center}
\includegraphics[width=0.48\columnwidth]{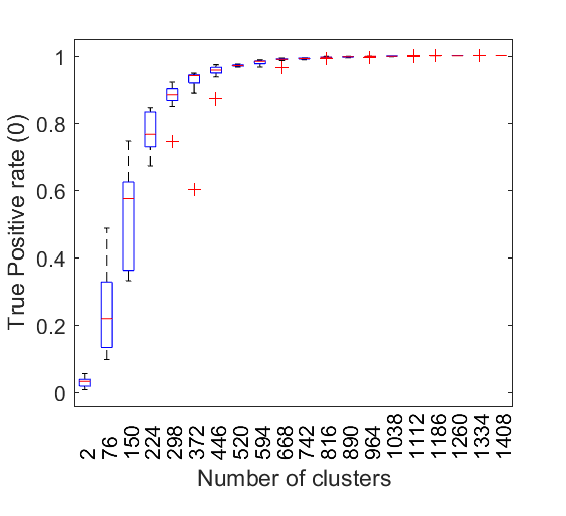}
\includegraphics[width=0.48\columnwidth]{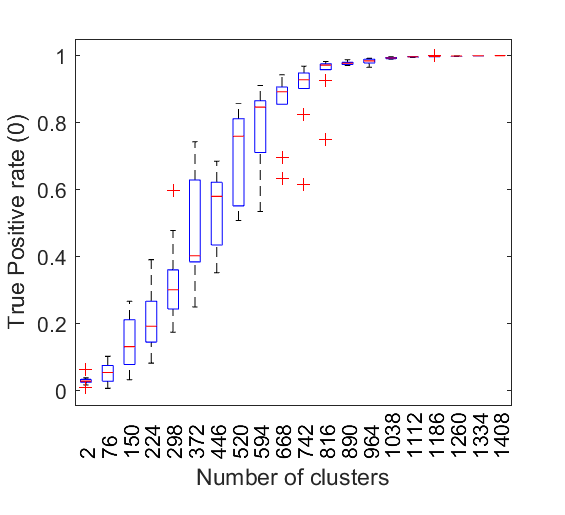}
\end{center}
\caption{True Positive rate at $\gamma=0$ for the combined system as a function of the number of clusters, $p$. Left panel corresponds to Algorithm \ref{alg:algorithm_1} {\it with} the optional projection step. Right panel corresponds to Algorithm \ref{alg:algorithm_1} {\it without} the projection step.}\label{fig:corrector_training}
\end{figure}

Before assessing performance of the ensemble, we evaluated filtering properties of the ensemble as a function of the number of clusters used. For consistency with predictions provided in Theorems \ref{thm:cube:1} -- \ref{thm:cube:k-separation},  Algorithm \ref{alg:algorithm_1} was used in this exercise.  Results of the test are shown in Fig \ref{fig:corrector_training}.
Note that as the number $p$ of clusters increases, the True Positive Rate at $\gamma=0$ as a function of $p$, approaches 1 regardless of the projection step. This is in agreement with theoretical predictions stemming from Theorem \ref{thm:cube:1-separation}. We also observed that performance drops rapidly with the average number of elements assigned to a cluster. In view of our earlier observation that vectors labelled as ``errors'' appear to be nearly orthogonal to each other, this drop is consistent with the bound provided in Corollary \ref{cor:k-separation}.

\begin{figure}[h]
\begin{center}
\includegraphics[width=0.45\columnwidth]{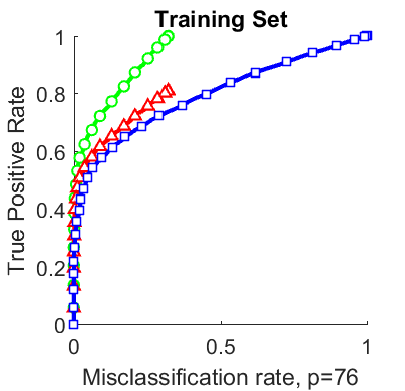}
\includegraphics[width=0.45\columnwidth]{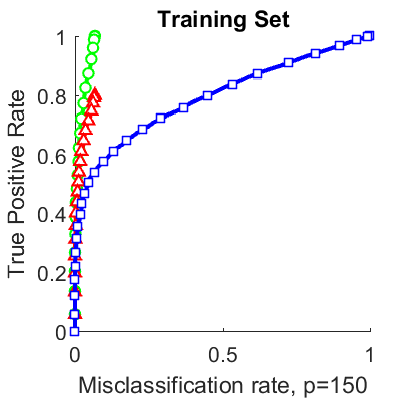}
\vspace{5mm}

\includegraphics[width=0.45\columnwidth]{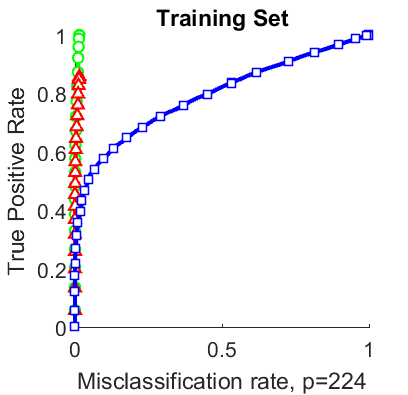}
\includegraphics[width=0.45\columnwidth]{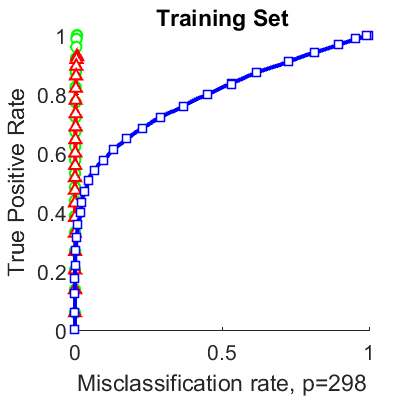}
\end{center}
\caption{Performance of the combined system after application of Algorithms \ref{alg:algorithm_1} (red triangles) and \ref{alg:algorithm_2} (green circles) on the training set for different numbers of clusters/nodes $p$ in step 5. Blue squares show performance of core AI system.}\label{fig:ensemble_training}
\end{figure}

\begin{figure}[h]
\begin{center}
\includegraphics[width=0.45\columnwidth]{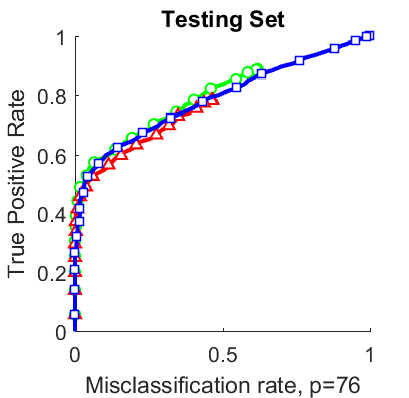}
\includegraphics[width=0.45\columnwidth]{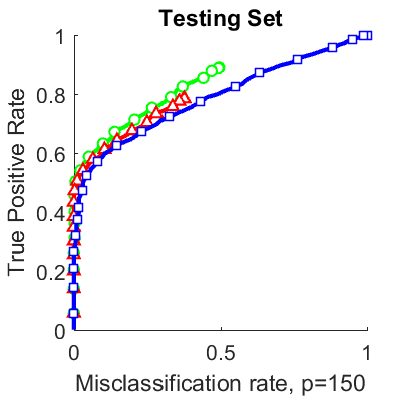}
\vspace{5mm}

\includegraphics[width=0.45\columnwidth]{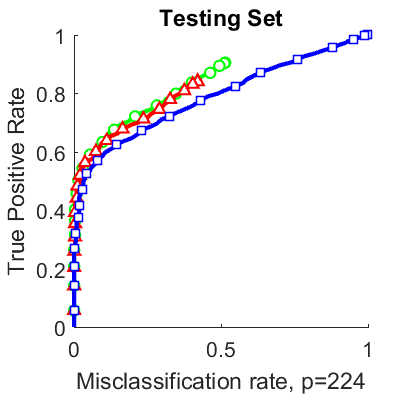}
\includegraphics[width=0.45\columnwidth]{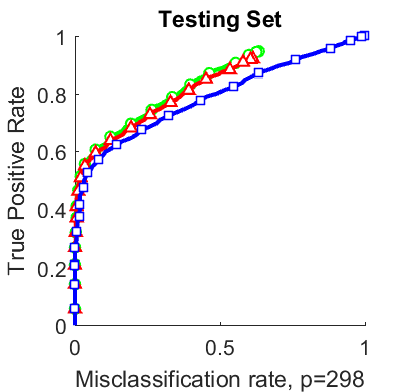}
\end{center}
\caption{Performance of the combined system after application of Algorithms \ref{alg:algorithm_1} (red triangles) and \ref{alg:algorithm_2} (green circles) on the testing set for different numbers of clusters/nodes $p$ in step 5. Blue squares show performance of core AI system.}\label{fig:ensemble_testing}
\end{figure}

Next  we assessed performance of Algorithms \ref{alg:algorithm_1}, \ref{alg:algorithm_2} and resulting ensembles on the training and testing sets for $p=76,150,224,298$. In both algorithms, optional projection step was used. The value of threshold $\theta$ was set to $0.2$. {In general, the value of threshold $\theta$ can be selected arbitrarily in  an interval of feasible values of $c_j$. When the optional projection step is used, this interval is $(0,1)$. Here we set the value of $\theta$ so that the hyperplanes $h_{j,2}$ in step 6 of Algorithm \ref{alg:algorithm_2} produced by standard perceptron algorithm  \cite{Rosenblatt1962}  consistently yielded perfect separation.}   Results are shown in Figs. \ref{fig:ensemble_training} and \ref{fig:ensemble_testing}.
According to Fig. \ref{fig:ensemble_training}, performance on the training set grows with $p$, as expected. This is aligned with theoretical results presented in Section \ref{sec:math_preliminaries}. The ensembles rapidly remove misclassification errors from the system, albeit at some cost to True Positive Rate. The latter cost for Algorithm \ref{alg:algorithm_2} appears to be negligible. On the testing set,  the picture changes. We see that, as the number $p$ of clusters/nodes grows, performance of the ensemble saturates, with Algorithm \ref{alg:algorithm_1} catching-up with Algorithm \ref{alg:algorithm_2}. This signals an over-fit and indicates a point where further increases in the number of nodes do not translate into expected improvements of the combined systems's performance. Apparent lack of correlations between feature vectors corresponding to errors (illustrated with Fig. \ref{fig:correlations_data}) may also contribute to the observed performance saturation on the testing set.

Notably, for $\gamma$ small both algorithms removed significant proportions of misclassification errors. This could be due to that training and testing sets for Algorithms \ref{alg:algorithm_1} and \ref{alg:algorithm_2} have been created from the performance analysis of the original core AI system at  $\gamma=0$.

\section{Conclusion}\label{sec:conclusion}

In this work we presented a novel technology for computationally efficient improvements of generic AI systems, including sophisticated Multilayer and Deep Learning neural networks.  These improvements are easy-to-train ensembles of elementary nodes. {After the ensembles are constructed, a possible next step could be} to  further optimize the system via e.g. error back-propagation. Mathematical operations at the nodes involve computations of inner {products followed} by standard non-linear operations like ReLU, step functions, {sigmoidal functions} etc. The technology was illustrated with a simple case study confirming its viability. {We note that the proposed ensembles can be employed for learning new tasks too.}


The proposed concept builds on our previous work \cite{GorTyuRom2016b} and complements results for equidistributions in a unit ball to product measure distributions. When the clustering structure is fixed, the method  is inherently one-shot and non-iterative, and its computational complexity scales at most linearly with the size of the training set. Sub-linear computational complexity of the ensembles construction makes the technology particularly suitable for large AI systems that have already been deployed and are in operation.

An intriguing and interesting feature of the approach is that training of the ensembles is largely achieved via Fisher linear discriminants. This, combined with the ideas from \cite{GGGMT2018prob}, paves {the way} for a potential relaxation of the i.i.d. assumption for the sampled data. Dealing with  this as well as exploring the proposed technology on a range of different AI architectures and applications is the focus of our future work.

\section{Acknowledgement}

The work is supported by Innovate UK Technology Strategy Board (Knowledge Transfer Partnership grants KTP009890 and KTP010522) and by  the Ministry of Education and Science of Russian Federation (Project No. 14.Y26.31.0022). The {authors} are thankful to Tatiana Tyukina for help and assistance with preparing the manuscript and artwork.

\section*{References}

\bibliography{extreme_cascade}

\appendix

\section{Proofs of theorems and other technical statements}\label{sec:appendix_proofs}

\subsection*{Proof of Theorem \ref{thm:cube:1}}

The proof follows immediately from Theorem 2 in \cite{GorbanTyukin:NN:2017} and Hoeffding inequality \cite{Hoeffding1963}. Indeed, if $t>0$, $X_i$ are independent bounded random variables, i.e.  $a_i\leq X_i\leq b_i$, $i=1,\dots,n$, and $\overline{X}=1/n \sum_{i=1}^n X_i$ then (Hoeffding inequality)
\[
\begin{split}
P\left(\overline{X}-E[\overline{X}] \geq t \right) \leq \exp\left(- \frac{2 n^2 t^2}{\sum_{i=1}^n(b_i-a_i)^2}\right)\\
P\left(|\overline{X}-E[\overline{X}]| \geq t \right) \leq 2\exp\left(- \frac{2 n^2 t^2}{\sum_{i=1}^n(b_i-a_i)^2}\right).
\end{split}
\]
Given that $x_{ij}=X_j$ where $X_j$ are independent random variables with $-1\leq X_j\leq 1$, $E[X_j^2]=\sigma_j^2$ (Assumption \ref{assume:product_measure}), we observe that $\|\bfx_i\|^2=\sum_{j=1}^n X_j^2$
\[
\begin{split}
&P\left(\left|\frac{\sum_{j=1}^n X_j^2}{n}-E\left[\frac{\sum_{j=1}^n X_j^2}{n}\right] \right|\geq t \right) =\\
&P\left(\left|\frac{\sum_{j=1}^n X_j^2}{n}-\frac{R_0^2}{n} \right| \geq t \right)=P\left(\left|\frac{\|\bfx_i\|^2}{R_0^2}-1\right| \geq \frac{tn}{R_0^2} \right)\\
&\leq 2 \exp\left(- 2 n t^2\right).
\end{split}
\]
Denoting $\varepsilon= {tn}/{R_0^2}$ and recalling that $0\leq X_j^2\leq 1$ we conclude that (\ref{eq:norms_cube}) holds true.

Noticing that $E[x_{ik} x_{jk}]=E[x_{ik}]E[x_{jk}]=0$, $E[y_k x_{ik}]=0$, $-1\leq x_{ik}\leq 1$, $-1\leq x_{jk}\leq 1$, and $-1\leq y_k x_{ik}\leq 1$, we observe that estimates (\ref{eq:dot_product_cube}) and (\ref{eq:dot_product_cube_fixed}) follow. $\square$

\subsection*{Proof of Theorem \ref{thm:cube:1-separation}}

Recall that  for any events $A_1,\dots,A_k$ the following estimate holds:
\begin{equation}\label{eq:elementary_probability}
P(A_1 \& A_2 \& \cdots \& A_k)\geq 1 - \sum_{i=1}^k(1-P(A_i)).
\end{equation}
Let
\[
 \frac{\|\bfx_{M+1}\|^2}{R_0^2}\geq 1-\varepsilon \ \ \mbox{(event} \ A_1 \mbox{)},
\]
and
\[
\left\langle \frac{\bfx_{M+1}}{R_0},\frac{\bfx_{i}}{R_0} \right\rangle < 1-\varepsilon \ \ \mbox{(events} \ A_2, \dots A_{M+1} \mbox{)}.
\]
The statement now follows from (\ref{eq:norms_cube}) and (\ref{eq:dot_product_cube}) with $\delta=1-\varepsilon$, and (\ref{eq:elementary_probability}). $\square$

\subsection*{Proof of Theorem \ref{thm:cube:k-separation}}

Suppose that $\|\bfx_i\|^2/R_0^2 \geq 1-\varepsilon$ for all $\bfx_i\in\mathcal{Y}$ (event $A_1$). Then
\[
\begin{split}
\ell_k(\bfx_i)=&\frac{1}{k}\left(\frac{\|\bfx_i\|^2}{R_0^2}+ \sum_{\bfx_j\in\mathcal{Y}, \ \bfx_j\neq \bfx_i} \left\langle\frac{\bfx_i}{R_0},\frac{\bfx_j}{R_0}\right\rangle \right)\geq \frac{1-\varepsilon +\beta (k-1)}{k}\\
&  \ \mbox{for all } \bfx_i\in\mathcal{Y}.
\end{split}
\]
Note that $\overline{\bfx}\in[-1,1]^n$ by construction. Consider events
\[
\ell_k(\bfx_j)=\left\langle\frac{\bfx_{j}}{R_0},\frac{\overline{\bfx}}{R_0}\right\rangle < \frac{1-\varepsilon + \beta (k-1)}{k} \ \mbox{(events} \ A_{1+j} \mbox{)}
\]
$j=1,\dots,M$. According to Theorem \ref{thm:cube:1} and (\ref{eq:elementary_probability})
\[
\begin{split}
& P(A_1)\geq 1 - k\exp \left(-\frac{2R_0^4 \varepsilon^2}{n} \right)\\
& P(A_{1+j})\geq 1 - \exp \left(-\frac{R_0^4 (1-\varepsilon+\beta(k-1))^2}{2k^2 n} \right).
\end{split}
\]
Hence
\[
\begin{split}
&P(A_1\&\cdots\& A_{1+M})\geq \\
& 1 - k\exp \left(-\frac{2R_0^4 \varepsilon^2}{n} \right) - M \exp \left(-\frac{R_0^4 (1-\varepsilon+\beta(k-1))^2}{2 k^2 n} \right).
\end{split}
\]
The result now follows $\square$.

\subsection*{Proof of Corollary \ref{cor:k-separation}}

Let $\|\bfx_i\|^2/R_0^2 \geq 1-\varepsilon$ for all $\bfx_i\in\mathcal{Y}$ (event $A_1$), and
\[
\left\langle\frac{\bfx_i}{R_0}, \frac{\bfx_j}{R_0}\right\rangle \geq - \delta  \ \mbox{for all} \ \bfx_i,\bfx_j\in\mathcal{Y}, \ \bfx_i\neq \bfx_j \ \mbox{(event } \ A_2 \mbox{)}.
\]
The corollary follows immediately from (\ref{eq:elementary_probability}) and Theorem \ref{thm:cube:1} (equations (\ref{eq:norms_cube}), (\ref{eq:dot_product_cube})) $\square$.

\subsection*{Proof of Corollary \ref{cor:cube:k-separation:2}}

 Let $\|\bfx_j\|^2\geq 1-\varepsilon$. Then $h(\bfx)\geq 0$ for all $\bfx\in\Omega$. Estimate (\ref{eq:k-separation:2}) hence follows from Theorem \ref{thm:cube:1} and  (\ref{eq:elementary_probability}). $\square$.

\subsection*{Proof of Lemma \ref{lem:cascade}}

The proof follows that of Theorem 3 in \cite{GorTyuRom2016b}. Let $\bfx$ be an element of $\mathcal{M}$ and $p_{h}$ be the probability of the event $h(\bfx)\geq 0$. Then the probability  of the event that $h(\bfx)\geq 0$ for at most $m$ elements from $\mathcal{M}$ is
\[
P(M,m)= \sum_{k=0}^{m-1} \left(\begin{array}{c}
                                    M\\
                                    k
                                 \end{array}\right) (1-p_h)^{M-k} p_h^k.
\]
Observe that  ${P}(M,m)$, as a function of $p_h$, is monotone and non-increasing on the interval $[0,1]$, with ${P}(M,m)=0$  at $p_h=1$ and ${P}(M,n)=1$ at $p_h=0$. Hence
\[
{P}(M,m) \geq \sum_{k=0}^{m-1} \left(\begin{array}{c}
                                    M\\
                                    k
                                 \end{array}\right) (1-p_{\ast})^{M-k} p_{\ast}^k = (1-p_\ast)^{M} \sum_{k=0}^{m-1} \left(\begin{array}{c}
                                    M\\
                                    k
                                 \end{array}\right) \left(\frac{p_{\ast}}{1-p_{\ast}}\right)^k.
\]
Given that  $\frac{(M-m+1)^k}{k!}\leq \left(\begin{array}{c}
                                    M\\
                                    k
                                 \end{array}\right)\leq \frac{M^k}{k!}$ for $ 0 \leq k\leq m-1$, we obtain
\[
{P}({M},m)\geq (1-p_\ast)^{M} \sum_{k=0}^{m-1} \frac{1}{k!}  \left(\frac{(M-m+1)p_{\ast}}{1-p_{\ast}}\right)^k.
\]
Expanding $e^x$ at $x=0$ with the Lagrange remainder term:
\[
e^{x}=\sum_{k=0}^{m-1} \frac{x^k}{k!} + \frac{x^m}{m!}e^{\xi}, \ \xi\in [0,x],
\]
results in the estimate
\[
\sum_{k=0}^{m-1} \frac{x^k}{k!}\geq e^x\left(1-\frac{x^m}{m!}\right) \ \mbox{for all} \ x\geq 0.
\]
Hence
\[
\begin{split}
&{P}({M},m)\geq\\
&(1-p_\ast)^{M} \exp\left(\frac{(M-m+1)p_{\ast}}{1-p_{\ast}}\right) \left(1-\frac{1}{m!}\left(\frac{(M-m+1)p_{\ast}}{(1-p_{\ast})}\right)^m\right).
\end{split}
\]
$\square$

\section{Hyperplanes vs ellipsoids for error isolation}\label{sec:planes_vs_ellipses}

Consider
\begin{equation}\label{eq:spherical_cap}
C_n(\varepsilon)= B_n(1)\cap \left\{\xi\in\Real^n \left|  \  \left\langle \frac{\bfx}{\|\bfx\|}, \xi \right\rangle  \geq 1-\varepsilon \right\} \right..
\end{equation}
Let
\[
{\rho}(\varepsilon)=(1-(1-\varepsilon)^2)^{\frac{1}{2}}.
\]
Note that ${\rho}(\varepsilon)$ is the radius of the ball containing the spherical cup $C_n$. Lemma \ref{lem:ball_vs_spherical_cap} estimates volumes of spherical caps $C_n(\varepsilon)$ relative to relevant $n$-balls of radius ${\rho}(\varepsilon)$.
\begin{lem}\label{lem:ball_vs_spherical_cap} Let  $C_n(\varepsilon)$ be a spherical cap defined as in (\ref{eq:spherical_cap}), $\varepsilon\in(0,1)$.  Then
\[
 \frac{\rho(\varepsilon)^{n+1}}{2} \left[\frac{1}{\pi^{\frac{1}{2}}} \frac{\Gamma\left(\frac{n}{2}+1\right)}{\Gamma\left(\frac{n}{2}+ \frac{3}{2}\right)}\right] < \frac{\Volume (C_n(\varepsilon))}{\Volume (B_n(1))} \leq \frac{\rho(\varepsilon)^n}{2}.
\]
\end{lem}

Note that \cite{ball1997elementary} $\Volume(B_n(r))=r^n \Volume(B_n(1))$ for all $n\in\Natural$ $r>0$. Hence the estimate of $\Volume (C_n(\varepsilon))$ from above is:
\begin{equation}\label{eq:cap_volume_estimate}
\Volume (C_n(\varepsilon))\leq \frac{1}{2}\Volume(B_{n}(1))\rho(\varepsilon)^{n}.
\end{equation}
Let us calculate the estimate of $\Volume (C_n(\varepsilon))$ from below. It is clear that
\[
\Volume (C_n(\varepsilon))=V(B_{n-1}(1))\int_{1-\varepsilon}^{1} (1-x^{2})^{\frac{n-1}{2}} dx
\]
The integral in the right-hand side of the above expression can be estimated from below as
\[
\begin{split}
&
\int_{1-\varepsilon}^{1} (1-x^{2})^{\frac{n-1}{2}} dx > \int_{1-\varepsilon}^{1} (1-x^{2})^{\frac{n-1}{2}} x dx \\
& = \frac{1}{2} \cdot \frac{1}{\frac{n}{2}+\frac{1}{2}}  \cdot (1-(1-\varepsilon)^2)^\frac{n+1}{2} = \frac{1}{2}  \cdot \frac{1}{\frac{n}{2}+\frac{1}{2}}    \cdot \rho(\varepsilon)^{n+1}
\end{split}
\]
Recall that $B_n(1)=\frac{\pi^\frac{n}{2}}{\Gamma\left(\frac{n}{2}+1\right)}$,  $\Gamma(x+1)=x\Gamma(x)$. Hence
\[
\begin{split}
&B_{n-1}(1)\cdot  \frac{1}{\frac{n}{2}+\frac{1}{2}} = \frac{\pi^\frac{n-1}{2}}{\Gamma\left(\frac{n}{2}+\frac{1}{2}\right)}\frac{1}{\frac{n}{2}+\frac{1}{2}}\\
&=\frac{\pi^\frac{n-1}{2}}{\Gamma\left(\frac{n+1}{2}+1\right)},
\end{split}
\]
and
\[
\int_{1-\varepsilon}^{1} (1-x^{2})^{\frac{n-1}{2}} dx > \frac{1}{2}\Volume(B_{n}(1))\rho(\varepsilon)^{n+1}\left[\frac{1}{\pi^{\frac{1}{2}}} \frac{\Gamma\left(\frac{n}{2}+1\right)}{\Gamma\left(\frac{n}{2}+ \frac{3}{2}\right)}\right]
\]
$\square$

\begin{cor}\label{cor:eBall_vs_eCap} Let  $C_n(\varepsilon)$ be a spherical cap defined as in (\ref{eq:spherical_cap}), $\varepsilon\in(0,1)$, and $B_n(k \rho(\varepsilon))$ be an $n$-ball with radius $k\rho(\varepsilon)$, $k\in \Real_{>0}$. Then
\[
\frac{\Volume (B_n(k\rho(\varepsilon)))}{\Volume (C_n(\varepsilon))} < k^{n} \frac{2\pi^{\frac{1}{2}}}{\rho(\varepsilon)} \left[\frac{\Gamma\left(\frac{n}{2}+1\right)}{\Gamma\left(\frac{n}{2}+ \frac{3}{2}\right)}\right]^{-1}
\]
\end{cor}

\begin{rem}
Using Stirling's approximation we observe that
\[
\frac{\Gamma\left(\frac{n}{2}+1\right)}{\Gamma\left(\frac{n}{2}+ \frac{3}{2}\right)}= O\left(n^{-\frac{1}{2}}\right).
\]
Thus $\frac{\Volume (B_n(\varepsilon))}{\Volume (C_n(\varepsilon))}<k^n  H(n,\varepsilon)$
where $H(n,\varepsilon)=O\left(\frac{n^{1/2}}{\rho(\varepsilon)}\right)$.
\end{rem}

According to Corollary \ref{cor:eBall_vs_eCap} the volumes of $B_n(\varepsilon)$, $B_n(\kappa \rho(\varepsilon))$, $\kappa\in(0,1)$  decay exponentially with dimension $n$ relative to that of $C_n(\varepsilon)$. This implies that distance-based detectors are extremely localized, and in comparison  with simple perceptrons, the proportion of points to which they respond positively is negligibly small.  On the other hand,  filtering properties of hyperplanes are extreme in high dimension (see Theorems \ref{thm:cube:1-separation}, \ref{thm:cube:k-separation} in Section \ref{sec:math_preliminaries}). This combination of properties makes perceptrons and their ensembles particularly attractive for fine-tuning of existing AI systems.

\end{document}